\documentclass[10pt,twocolumn,letterpaper]{article}

\usepackage[pagenumbers]{cvpr2023/cvpr} 

\usepackage{graphicx}
\usepackage{amsmath}
\usepackage{amssymb}
\usepackage{booktabs}
\usepackage{algorithm,algorithmic,bm,color}

\usepackage[pagebackref,breaklinks,colorlinks]{hyperref}

\usepackage[capitalize]{cleveref}
\crefname{section}{Sec.}{Secs.}
\Crefname{section}{Section}{Sections}
\Crefname{table}{Table}{Tables}
\crefname{table}{Tab.}{Tabs.}

\def\cvprPaperID{2202} 
\def\confName{CVPR}
\def\confYear{2023}

\usepackage{url}
\usepackage{subfiles}
\usepackage{graphicx}
\usepackage{term_to_use}
\usepackage{boldline}
\usepackage{enumitem}
\usepackage{multirow}
\usepackage{bbm}
\usepackage{setspace}

\begin{document}

\title{\approach{}: Discriminative Geometry-Aware Learning for\\ Generalized Few-Shot Object Detection}

\author{Jiawei Ma \quad Yulei Niu$^{*}$ \quad Jincheng Xu \quad Shiyuan Huang \quad Guangxing Han \quad Shih-Fu Chang\\
Columbia University\\
{\tt\small \{jiawei.m, yulei.niu, jx2467, shiyuan.h, gh2561, sc250\}@columbia.edu}
}
\maketitle
\let\thefootnote\relax\footnotetext{$^{*}$Corresponding Author.}

\begin{abstract}
    Generalized few-shot object detection aims to achieve precise detection on both base classes with abundant annotations and novel classes with limited training data. Existing approaches enhance few-shot generalization with the sacrifice of base-class performance, or maintain high precision in base-class detection with limited improvement in novel-class adaptation. In this paper, we point out the reason is insufficient
    \underline{Di}scriminative feature learning for all of the classes. As such, we propose a new training framework, DiGeo, to learn \underline{Geo}metry-aware features of inter-class separation and intra-class compactness. To guide the separation of feature clusters, we derive an offline simplex equiangular tight frame (ETF) classifier whose weights serve as class centers and are maximally and equally separated. To tighten the cluster for each class, we include adaptive class-specific margins into the classification loss and encourage the features close to the class centers. Experimental studies on two few-shot benchmark datasets (VOC, COCO) and one long-tail dataset (LVIS) demonstrate that, with a single model, our method can effectively improve generalization on novel classes without hurting the detection of base classes. Our code can be found \href{https://github.com/Phoenix-V/DiGeo}{here}. 
\end{abstract}


\section{Introduction}\label{sec:intro}

Recent years have witnessed the tremendous growth of object detection through deep neural models and large-scale training~\cite{ren2015faster,carion2020end,redmon2016you,zhu2020deformable,zhang2022dino,sun2021sparse,R_RPN,han2018semi,SSD_TDR}. However, the success of detection models heavily relies on the amount and quality of annotations, which requires expensive annotation cost and time. In addition, traditional object detection models perform worse on the classes with a limited number of annotations~\cite{wang2020frustratingly,Han_2022_CVPR,yan2019meta}, while human are able to learn from few observations. In order to close the gap between human vision system and detection models, recent studies have investigated how to generalize well on rare classes under the few-shot object detection (FSOD) setting.
Specifically, given many-shot (\textit{base}) classes with plenty of training data and few-shot (\textit{novel}) classes with extremely limited training data (\eg, 5 annotated instances per class), FSOD expects the model to detect the objects in the novel classes well.

\begin{figure}
    \centering
    \includegraphics[width=0.95\linewidth]{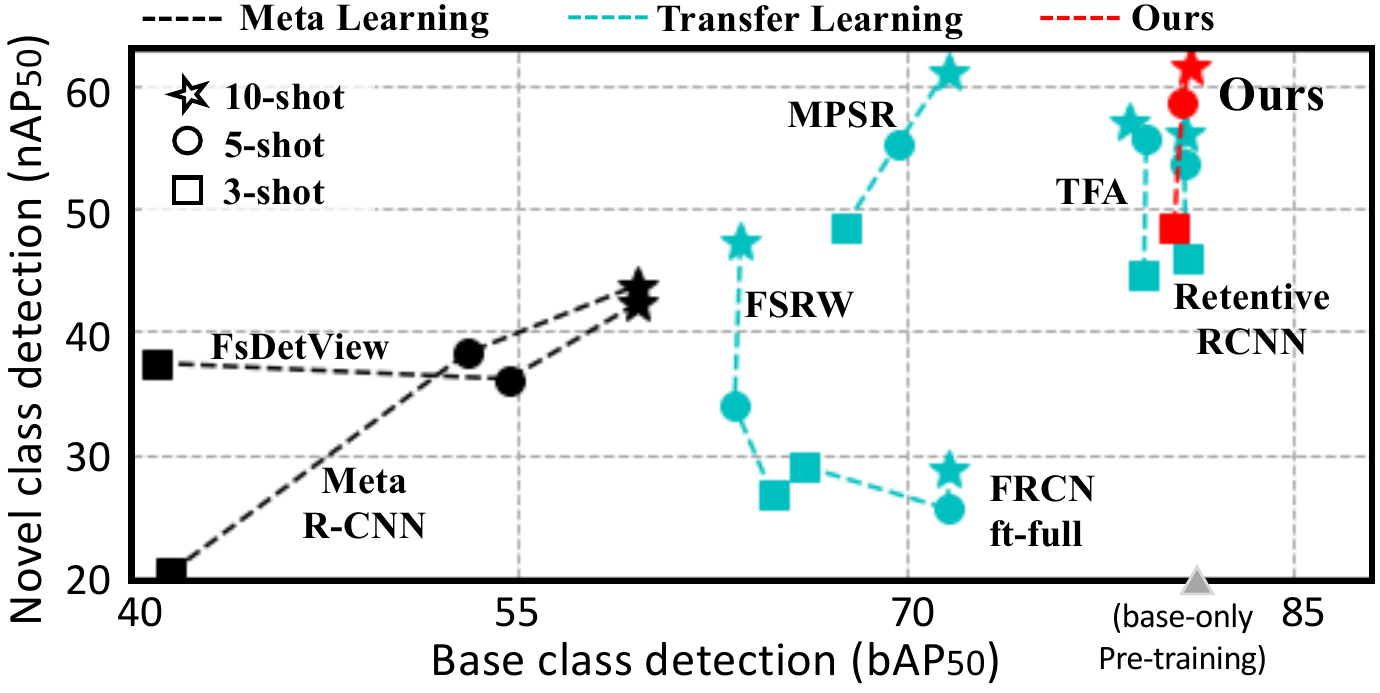}
    \caption{Performance on few-shot object detection on Pascal VOC~\cite{pascal-voc-2007}. 
    Previous transfer-learning approaches (\textcolor{cyan}{blue}) balancing the training data by aggressively down-sampling the base set and may result in overfitting. Instead, we (\textcolor{red}{red}) use the full train set, aiming to both maintain precise base detection but learn discriminative features from the limited annotations for few-shot classes.
    }
    \label{fig:concept}
\end{figure}

To improve the generalization ability on novel-class detection, recent studies~\cite{sun2021fsce,wang2020frustratingly,fan2021generalized} conduct transfer learning in a two-step manner. In detail, the model is pre-trained on the whole set of base classes, and then fine-tuned on the union of the set of novel classes and an aggressively down-sampled base subset.
However, the efficient few-shot adaptation is often achieved at the expense of sacrificing precision on base detection (Fig.~\ref{fig:concept}).
Being aware of this limitation, Fan \etal~\cite{fan2021generalized} proposed to evaluate the performance of both base and novel classes in the generalized few-shot object detection (GFSOD) setting. In addition, they proposed a consistency regularization to emphasize the pre-trained base knowledge during fine-tuning and employed an ensembling strategy.
However, they design different classifiers for base and novel classes, and the adaptation on novel classes is impeded due to a complex ensembling process.

In this paper, we pointed out that the devil is in insufficient discriminative feature learning for few-shot object detection, including inefficient knowledge adaptation to novel classes and unexpected knowledge forgetting of base classes. First, as the novel instances are extremely limited during training, it is hard to capture the representative visual information of novel classes and adapt the knowledge learned from base classes to novel classes.
As a result, the model cannot distinguish between the novel classes, which weakens the few-shot adaptation. Secondly, balanced training strategies such as down-sampling fail to utilize the diverse training samples from base set. Thus, it is hard to preserve the complete knowledge of base classes, which leads to overfitting and further decreases the detection scores.

To tackle these challenges, we proposed a new training framework, \textbf{\approach{}}, to make the best of both worlds for generalized few-shot object detection, \ie, improving generalization on novel classes without hurting the detection of base classes. Our motivation is to learn \textbf{Di}scriminative \textbf{Geo}metry-aware features via \textit{inter-class separation} and \textit{intra-class compactness}. For inter-class separation, we expect the class centers~\cite{wen2016discriminative} to be well distinct from each other. 
Motivated by the symmetric geometry of simplex equiangular tight frame (ETF)~\cite{papyan2020prevalence}, we proposed to use ETF as classifier to guide the separation of features. To be specific, we derive an offline ETF whose weights are maximally \& equivalently separated (\ie, independent from the training data distribution) and are assigned as fixed centers for all classes.
For intra-class compactness, we expect the features to be closed to the class centers for a clear decision boundary. 
In practice, we add class-specific margins to output logits during training to 
push the features close to the class centers.
The margins are based on instance distribution prior and are then adaptively adjusted though self-distillation. Meanwhile, we consider the huge imbalance between \textit{base} set and \textit{novel} set, and up-sample the \textit{novel} set to facilitate the feature extraction.

We validate the effectiveness of \approach{} under the GFSOD setting on Pascal VOC~\cite{pascal-voc-2007,pascal-voc-2012} and MS COCO~\cite{lin2014microsoft}.
Compared to existing methods, we can both achieve precise detection on base classes and sufficiently improve the adaptation efficiency on novel classes using a single model. 
Furthermore, our \approach{} can be intuitively extended to long-tailed object detection. Experimental results on LVIS datasets demonstrate the generalizibility of our approach.
Our contributions are summarized as follows:
\vspace{-2.5mm}
\begin{itemize}[leftmargin=*]
    \item We revisit few-shot object detection from a perspective of discriminative feature learning, and point out that existing methods fail in knowledge adaptation to novel classes and suffer from knowledge forgetting of base classes.
    \vspace{-2.5mm}
    \item We propose \approach{} to pursue an desired feature geometry, \ie, inter-class separation and intra-class compactness, which consistently improves the performance on both base and novel classes.
    \vspace{-2.5mm}
    \item We conduct extensive experiments on three benchmark datasets for few-shot object detection and long-tailed object detection to verify the generalizability of \approach{}.
\end{itemize}

\section{Related Work}

\begin{figure*}[]
    \centering
    \includegraphics[width=0.95\linewidth]{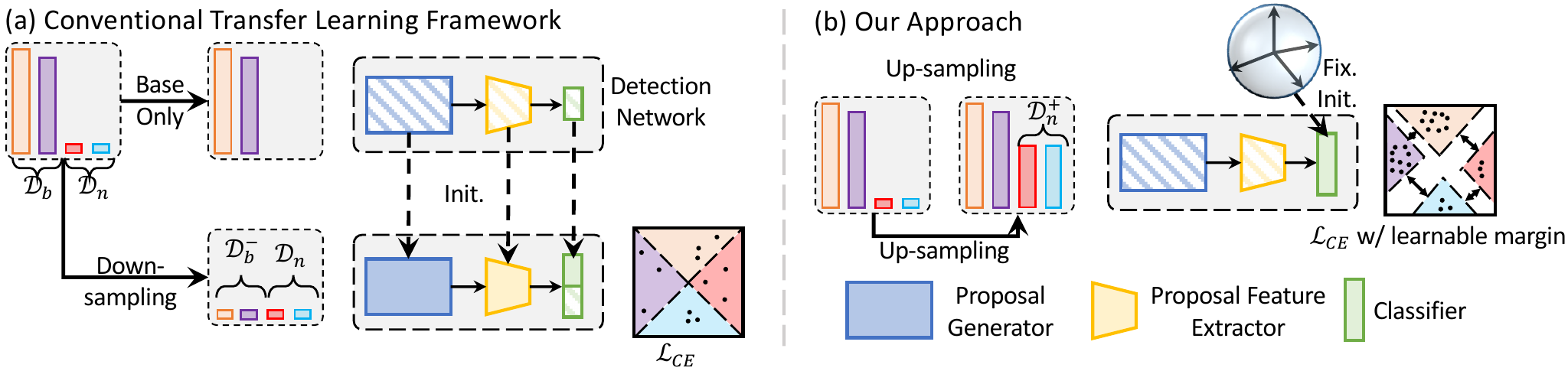}
    \caption{Comparison of training frameworks. (a) Conventional approaches first pre-train a base detector among the \textit{base} set $\baseset{}$ and then finetune on the union of novel set $\novelset{}$ and down-sampled subset of base classes $\basesubset{}$. (b) Instead, we choose to up-sample $\novelset{}$ and directly train the detector on the full set. We derive a fixed classifier offline with maximally \& equally separated weights and learn the adaptive margins to tighten the feature clusters. The margins are estimated from priors of instance distribution and learned though self-distillation. The block with shading means training from scratch. We use the same design for localization and omit it for simplicity.}
    \label{fig:appraoch}
\end{figure*}

\bitem{Few-shot object detection (FSOD)} aims to detect objects of few-shot (\textit{novel}) classes at instance-level. 
To improve the adaptation efficiency, the approaches based on the meta-learning and the transfer-learning are investigated.
The \emph{meta-learning approaches \cite{fan2020few,han2022meta,Han_2022_CVPR,kang2019few,Han_2021_ICCV,han2022multimodal}} learns a class-agnostic meta-learner to align instances of the same class from different images. 
Under the Faster-RCNN framework, the attention-based meta-RPN~\cite{fan2020few} and meta-detector~\cite{han2022meta} are proposed to generate class-relevant proposals and improve the instance alignment.
In addition, approaches based on Transformer~\cite{Han_2022_CVPR} and YoLo~\cite{kang2019few} are proposed to extract features jointly and align features at multiple scales. 
The \textit{transfer-learning approaches \cite{wang2020frustratingly,wu2020multi,sun2021fsce,ma2022few}} performs finetuning for few-shot adaptation. 
Specifically, TFA~\cite{wang2020frustratingly} pre-trains an base detector from plenty of \textit{base} samples and finetune it for novel classes. 
To improve the adaptation efficiency, multi-scale feature extraction~\cite{wu2020multi} and regularization such as contrastive loss~\cite{sun2021fsce}, margin equilibrium~\cite{li2021beyond} and transformation invariance~\cite{li2021transformation} are employed.
Recently, DeFRCN~\cite{qiao2021defrcn} adjusts gradients back-propagated from different losses and achieve superior novel detection scores. 

\bitem{Generalized Few-Shot Object Detection.}
For all FSOD approaches mentioned above, the precision on \textit{base} detection is sacrificed after few-shot adaptation. 
This phenomenon has also been observed in various vision tasks where models forget the base knowledge due to domain gap or distribution gap~\cite{tang2020unbiased,niu2021counterfactual,qi2020two,niu2021introspective,zhu2022cross,zhu2022prompt,niu2022respecting}.
As pointed out by Fan \etal~\cite{fan2021generalized}, different from the classification~\cite{finn2017model,snell2017prototypical,tian2020rethinking,ma2021partner,zhao2021domain,liu2020negative,Han_2023_CVPR,huang2022task,ypsilantis2021met},
an image may contain instances from both novel and base classes and base detection is also important.
Then, they propose a consistency regularization few-shot fine-tuning and employ an model ensembling technique
to preserve the precision of base detection. However, the few-shot adaptation efficiency is inevitably limited.
In a more general case, \emph{long-tail object detection} (LTOD) has been studied where techniques such as resampling~\cite{zhang2021distribution,ren2020balanced}, decoupling~\cite{li2020overcoming,wang2020devil} and reweighting~\cite{zhang2021distribution,li2022equalized} are studied. Also, ACSL~\cite{wang2021adaptive} revisits LTOD from a statistic-free perspective and propose the adaptive suppression loss.

\bitem{Feature Distribution on a Balanced Set} has been studied in classification. To be specific, the weights in the last linear layer is treated as class centers where the geometry property of feature output by pernuminate layer is analyzed. Recently, Papyan \etal~\cite{papyan2020prevalence} summarized it as neural collapse (NC) and observed that 1) the features in the same class are maximally concentrated towards the class mean and different feature clusters are maximally separated~\cite{yang2022we}. 2) The class means and the class centers converge to each other.  

\section{Background}

We first introduce the few-shot object detection (FSOD) task, and analyze the limitations of existing FSOD methods.

\subsection{Few-shot Object Detection}\label{sec:task-pre}

In this paper, we focus on the task of few-shot object detection (FSOD). The training data consists of a \textit{base} set $\baseset{}$ and a \textit{novel} set $\novelset{}$, where the \textit{base} classes $\basecls{}$ have plenty of annotated object instances while \textit{novel} classes $\novelcls{}$ has limited annotations.
In an $N_n$-way $K$-shot FSOD task with $|\novelcls{}|=N_n$, each novel class has $K$ annotated instances.  
Note that an image may contain multiple instances from different classes with associated bounding boxes, which is more challenging than the few-shot classification where each image contains one object to be recognized.
Then, we follow ~\cite{fan2021generalized,wang2020frustratingly} to validate the robustness of detection model under the generalized few-shot object detection (GFSOD) setting, where the test samples come from both base and novel classes, and the models are evaluated on all classes.

Commonly, object detection models consist of a proposal generation module to generate a set of region candidates, and a detection module to localize \& classify objects on the extracted proposals~\cite{ren2015faster,redmon2016you,carion2020end,zhu2020deformable,sun2021sparse}. 
For the classification part, an additional background class should be considered to recognize the proposal with no foreground objects. We regard the last linear layer as classifier, and its weights $W = \{\mathbf{w}_i\}_{i=1}^{N_b+N_n+1}$ as class centers where $N_b=|\mathcal{C}_b|$. 
Without loss of generality, we set $W_b = \{\mathbf{w}_i\}_{i=1}^{N_b}$, $W_n = \{\mathbf{w}_i\}_{i=N_b+1}^{N_b+N_n}$, and $W_- = \{\mathbf{w}_{N+1}\}$ as weights for base classes $\basecls{}$, novel classes $\novelcls{}$, and  background $\negcls{}$.

\subsection{Analysis of Existing Methods}\label{sec:preliminary}

As a representative \textit{transfer-learning} approach shown in Fig.~\ref{fig:appraoch}(a), TFA~\cite{wang2020frustratingly} first trains a \textit{Base} detector on $\baseset{}$ for $\basecls{}$ as initialization. Then, in an $N_n$-way $K$-shot GFSOD task, $K$ instances for each \textit{base} class $c \in \basecls{}$ from $\baseset{}$ are selected to a subset $\basesubset{}$. The detector is fine-tuned on  $\basesubset{} \cup \novelset{}$ with balanced training data distribution over $\basecls{} \cup \novelcls{}$. However, for each class, as the training data is extremely limited, overfitting to $\basesubset{}$ is unignorable and results in the drop of base detection. As such, Retentive RCNN~\cite{fan2021generalized} proposes to ensemble the detector adapted for $\basecls{} \cup \novelcls{}$ and the \textit{Base} detector by combining their outputs as final prediction. However, the novel detection performance on $\novelcls{}$ is limited. 

Nevertheless, training among $\baseset{} \cup \novelset{}$ makes the model favor $\basecls{}$. As shown in Fig.~\ref{fig:step-center-ablation}(a), the novel weights $W_n$ are not well-learned and close to weights of other foreground classes. 
%
With such a classifier, the proposal features (\ie, input feature of classifier) cannot be separated. Thus, as shown in Fig.~\ref{fig:appraoch}(b), we obtain a classifier offline with well-separated weights. For each class, the features are trained to be compact and close to the centers using learnable margins.

\section{Approach}

Considering the limitations mentioned above, we aim to achieve the best of both worlds using a single model, \ie, improve the few-shot adaptation performance on novel classes without hurting the precision on base detection. Our motivation is to enhance the discriminative feature learning of detection models, \ie, clear boundaries on the feature space to discriminate all classes. We realize this idea from two aspects, inter-class separation between all classes and intra-class compactness for each class. 

\subsection{Inter-Class Separation}\label{sec:inter}

We realize inter-class separation by maximizing the pair-wise distances between class centers. Specifically, for each $\mathbf{w}_i$, we maximize its minimum 
distance with all other weights $W\setminus\{\mathbf{w}_i\}$:
\begin{equation}
\begin{gathered}
    W^* = \textrm{argmax}_{W}\sum\nolimits^{N_c}_{i=1} \min\nolimits_{j,i \neq j} \|\mathbf{w}_i-\mathbf{w}_j\|_2^2 \\
    s.t.~\|\mathbf{w}_i\|=1,~~\forall \mathbf{w}_i \in W
\end{gathered}\label{eq:etf}
\end{equation}
where $N_c=N_b+N_n+1$ and all weight vectors are of the same norm (\eg, 1). When the feature dimension $d \geq N_c-1$, the distances of all class center pairs in $W^*$ should be the same. Also, the angle between any two of the class centers has the same value given $\|\mathbf{w}_i\|=1$. In this way, we expect the class centers to be evenly distributed in the feature space. In this case, $W^*$ is equivalent to simplex equiangular tight frame (ETF)~\cite{papyan2020prevalence}. Furthermore, we have the following theorem for ETF.

\noindent\textbf{Theorem} \textit{Suppose the vector space is $d$-dimensional and the number of vectors is $N$. When $d \geq N-1$, we can always derive a simplex ETF whose vectors are maximally and equally separated from each other.}

The above theorem guarantees the existence of ETF in application when $d \geq N-1$. For $d < N_c-1$, \eg, the number of classes is large while the feature dimension is compact, we can project the $d$-dim feature to a $d'$-space space with $d' \geq N_c-1$. Then, we can always obtain a Simplex ETF classifier in the mapped feature space.

We have two options to obtain the Simplex ETF classifier. The \emph{online} solution is to use Eq.~\eqref{eq:etf} as a regularization loss to learn the classifier during training. The \emph{offline} solution is to manually set the Simplex ETF classifier for all classes $\mathcal{C} = \basecls{} \cup \novelcls{} \cup \{\negcls{}\}$ and fix it during training. We experimentally find the offline solution is more stable and better than the online solution (details discussed in Sec.~\ref{sec:discussion}) and thus use the offline solution in implementation.

\subsection{Intra-Class Compactness}\label{sec:intra}

We realize intra-class compactness by tightening the clusters of features and push the samples close to the assigned center in $W^{*}$. The challenges are two folds. First, the number of training samples in base and novel classes extremely are imbalanced, which makes it hard to determine the boundaries of novel classes in the feature space. Second, as the number of novel classes is much smaller than that of base classes, \ie, $|\novelset{}| \ll |\baseset{}|$, the network receives less positive gradients for novel classes~\cite{wang2021adaptive}, which makes the features of instances in novel classes farther to the class centers and thus less discriminative. 

Inspired by the success of logit adjustment in long-tailed recognition~\cite{menon2020long}, we apply class-specific margins on logits to modify the classification loss and balance the optimization between base and novel classes. Specifically, we calculate the class-specific margins based on the frequencies of instance (\ie, bounding box annotations) as priors:

\begin{equation}
    m_c=
    \begin{cases}
    -\log(p_c) & \text{, if $c\in\basecls{} \cup \novelcls{}$}\\
    -\log(p_-) & \text{, if $c=\negcls{}$}\\
    \end{cases},
\end{equation}
where $p_c$ is the frequency of bounding box annotations for class $c$, and $p_-$ is an estimated probability of background boxes to train the classifier, and $p_- + \sum_{c\in\basecls{} \cup \novelcls{}}p_c=1$.  Intuitively, the class with fewer data is assigned with a larger margin to guarantee the learning of this class.

Suppose that the logit outputs for sample $x$ are $\mathbf{v} = \{v_c\}_{c \in \mathcal{C}}$, we use the following prior-margin cross-entropy loss by adding the margins to the logits:
\begin{equation}\label{eq:prior}
    \mathcal{L}_\text{prior}(x) = - \sum_{c\in\mathcal{C}}y_c\cdot\log \frac{\exp(v_c - m_c)}{\sum_{c' \in \mathcal{C}}\exp(v_{c'} - m_{c'})}.
\end{equation}
where $y_c$ equals to 1 if $c$ is the ground-truth label, otherwise $y_c=0$. Note that our prior-margin loss reduce to vanilla cross-entropy loss if all margins $m_c$ are set as 0. As the margins are obtained based on prior distribution and fixed during training, we term this baseline as \fixstage{}. 

\begin{table*}[]
\centering
    \caption{Performance comparison of AP$_{50}$ on the PASCAL VOC dataset on all classes $\basecls{} \cup \novelcls{}$. The \best{best} and \subest{second-best} are highlighted.}
    \resizebox{\linewidth}{!}
    {\renewcommand{\arraystretch}{1.05}
    \begin{tabular}{l|ccccc|ccccc|ccccc|c}
    \hlineB{3}
        \multirow{2}{*}{Approach} & \multicolumn{5}{c|}{split 1}      & \multicolumn{5}{c|}{split 2}      & \multicolumn{5}{c|}{split 3}  & \multirow{2}{*}{Avg}    \\ 
                                   & 1    & 2    & 3    & 5    & 10   & 1    & 2    & 3    & 5    & 10   & 1    & 2    & 3    & 5    & 10 &  \\ \hline
                                   \multicolumn{17}{c}{Meta-Learning Approaches} \\ \hline
        Meta RCNN~\cite{yan2019meta}$^{*}$      & 17.5 & 30.5 & 36.2 & 49.3 & 55.6 & 19.4 & 33.2 & 34.8 & 44.4 & 53.9 & 20.3 & 3.0  & 41.2 & 48.0 & 55.1 & 38.0 \\
        FSRW~\cite{perez2020incremental}        & 53.5 & 50.2 & 55.3 & 56.0 & 59.5 & 55.1 & 54.2 & 55.2 & 57.5 & 58.9 & 54.2 & 53.5 & 54.7 & 58.6 & 57.6 & 55.6\\
        FsDetView~\cite{xiao2020few}$^{*}$      & 36.4 & 40.3 & 40.1 & 50.0 & 55.3 & 36.3 & 43.7 & 41.6 & 45.8 & 54.1 & 37.0 & 39.5 & 40.7 & 50.7 & 54.8 & 44.4 \\ \hline
        \multicolumn{17}{c}{Transfer-Learning Approaches} \\ \hline
        TFA w/ fc~\cite{wang2020frustratingly}  & 69.3 & 66.9 & 70.3 & 73.4 & 73.2 & 64.7 & 66.3 & 67.7 & 68.3 & 68.7 & 67.8 & 68.9 & 70.8 & 72.3 & 72.2 & 69.5\\
        TFA w/ cos~\cite{wang2020frustratingly} & \subest{69.7} & 68.2 & 70.5 & 73.4 & 72.8 & 65.5 & 65.0 & 67.7 & 68.0 & 68.6 & 67.9 & 68.6 & 71.0 & 72.5 & 72.4 & 69.5 \\
        FRCN-ft-full~\cite{yan2019meta}$^{*}$   & 55.4 & 57.1 & 56.8 & 60.1 & 60.9 & 50.1 & 53.7 & 53.6 & 55.9 & 55.5 & 58.5 & 59.1 & 58.7 & 61.8 & 60.8 & 57.2\\
        MPSR~\cite{wu2020multi}                 & 56.8 & 60.4 & 62.8 & 66.1 & 69.0 & 53.1 & 57.6 & 62.8 & 64.2 & 66.3 & 55.2 & 59.8 & 62.7 & 66.9 & 67.7 & 62.1\\
        Retentive R-CNN~\cite{fan2021generalized}$^{\dagger}$& \best{71.3} & \best{72.3} & \subest{72.1} & \subest{74.0} & \subest{74.6} & \subest{66.8} & \best{68.4} & \subest{70.2} & \subest{70.7} & \subest{71.5} & \best{69.0} & \best{70.9} & \subest{72.3} & \subest{73.9} & \subest{74.1} & \subest{71.5} \\ \hline 
        \approach{} (Ours) & \subest{69.7} & \subest{70.6} & \best{72.4} & \best{75.4} & \best{76.1} & \best{67.5} & \best{68.4} & \best{71.4} & \best{71.6} & \best{73.6} & \subest{68.6} & \best{70.9} & \best{72.9} & \best{74.4} & \best{75.0} & \best{71.9}  \\
    \hlineB{3}
    \multicolumn{16}{l}{$^{*}$: results reported by Retentive R-CNN~\cite{fan2021generalized} and TFA~\cite{wang2020frustratingly}.$^{\dagger}$: Model ensembling.}\\
    \end{tabular}
    }\label{tab:voc-all}
\end{table*}

\begin{table}[]
\centering
    \caption{Comparison of nAP$_{50}$ and bAP$_{50}$ on the PASCAL VOC.}
    \resizebox{\linewidth}{!}
    {\renewcommand{\arraystretch}{1.15}
    \begin{tabular}{l|ccccc|c}
    \hlineB{3}
        \multirow{2}{*}{Approach} & \multicolumn{5}{c|}{nAP$_{50}$ (Avg. on splits for each shot)}      & bAP$_{50}$ \\ 
                                   & 1    & 2    & 3    & 5    & 10   &  Avg. \\ \hline
        Meta R-CNN~\cite{yan2019meta}$^{*}$     & 11.2 & 15.3 & 20.5 & 29.8 & 37.0 & 43.1 \\
        FSRW~\cite{perez2020incremental}        & 16.6 & 17.5 & 25.0 & 34.9 & 42.6 & 65.0 \\
        FsDetView~\cite{xiao2020few}$^{*}$      & 26.9 & 20.4 & 29.9 & 31.6 & 37.1 & 49.5 \\ \hline
        TFA w/ fc~\cite{wang2020frustratingly}      & 27.6 & 30.6 & 39.8 & 46.6 & 48.7 & 79.6 \\
        TFA w/ cos~\cite{wang2020frustratingly}     & 31.4 & 32.6 & 40.5 & 46.8 & 48.3 & 79.3 \\
        FRCN-ft-full~\cite{yan2019meta}$^{*}$   & 16.1 & 20.6 & 28.8 & 33.4 & 36.5 & 67.2 \\
        MPSR~\cite{wu2020multi}           & \best{36.2} & \best{37.2} & \subest{44.6} & \subest{49.1} & \subest{53.2} & 68.1 \\
        Retentive RCNN~\cite{fan2021generalized}$^{\dagger}$ & 31.4 & \subest{37.1} & 41.4 & 46.8 & 48.8 & \best{81.6} \\ \hline
        \approach{} (Ours) & \subest{31.6} & 36.1 & \best{45.8} & \best{51.2} & \best{55.1} & \subest{81.3} \\
    \hlineB{3}
    \multicolumn{7}{l}{$^{*}$: results are reported by Retentive R-CNN~\cite{fan2021generalized} and TFA~\cite{wang2020frustratingly}.}\\
    \multicolumn{7}{l}{$^{\dagger}$: Model ensembling. Full tables can be found in Supp.} \\
    \end{tabular}
    }\label{tab:voc-all-detail}
\end{table}

Though the margin-based loss is calculated over all the proposals, precisely calculating the margins from the proposals is time-consuming. Thus, we obtain the prior margins over all annotated bounding box instances. In this case, there is a misalignment between proposal-based loss and instance-based margin. To mitigate this gap, we proposed to adaptively learn the margins based on the priors. Motivated by the success of self distillation~\cite{tian2020rethinking} in knowledge transfer, we use the detection module learned from $\mathcal{L}_\text{prior}$ in Eq.~\eqref{eq:prior} as teacher model, and distill its knowledge to a student model to adaptively learn and update the margins through soft labels, which has the same architecture as teacher model but different parameters. For sample $x$, the ground-truth label is $y$, the adaptive-margin distillation objective is:
\begin{equation}
    \mathcal{L}_\text{adapt}(x) = - \sum_{c \in \mathcal{C}} \frac{p_c^{t} + y_c}{2} \log \frac{\exp(v^s_c - m_c^{s})}{\sum_{c' \in \mathcal{C}}\exp(v^s_{c'} - m_{c'}^{s})},
\end{equation}
where the predicted probability for class $c$ of the teacher model $p^t_c$ is obtained by $\frac{\exp(v_c - m_c)}{\sum_{c' \in \mathcal{C}}\exp(v_{c'} - m_{c'})}$, $v^s_c$ denotes the logit output for class $c$ of the student model, and $m^s_c$ denotes the adaptive learnable margin for class $c$. The teacher model is fixed during self distillation, and the student detection head uses the same ETF classifier weights $W^{*}$ with other parts in the detection module to be learned. Finally, we use the student model for evaluation.

Even though the margins are added during training, the extreme imbalance between base set and novel set still makes the detector favors more on base set. Considering this limitation and the challenge that the number of novel classes is very limited to provide the gradients for network updating, we proposed to up-sample the images containing annotations of novel classes ($\mathcal{D}^+_n$).  Specifically, we use repeated factor sampling (RFS)~\cite{gupta2019lvis} and the repeating times is set by a hyper-parameter threshold in RFS. We experimentally found that using up-sampling itself can achieve marginal improvement, but can clearly improve the novel detection precision combined with our approach. This observation demonstrates that the up-sampling strategy works closely with our hypothesis rather than just a trivial trick.

\section{Experiment}

We mainly conduct experiments on the few-shot object detection (FSOD) benchmark datasets Pascal VOC and MS COCO to validate the effectiveness of our proposed~\approach{}. We further apply~\approach{} on long-tailed object detection and conduct experiments on LVIS to show its generalizability.


\begin{table*}[]
\centering
    \caption{Performance comparison of MS COCO dataset.}
    \resizebox{\linewidth}{!}
    {\renewcommand{\arraystretch}{1.12}
    \begin{tabular}{l|cccccccc|cccccccc}
    \hlineB{3}
        \multirow{2}{*}{Approach} & \multicolumn{8}{c|}{10-shot} & \multicolumn{8}{c}{30-shot} \\
         & AP   & bAP  & nAP  & nAP$_{50}$ & nAP$_{75}$ & nAPs & nAPm & nAPl & AP   & bAP  & nAP  & nAP$_{50}$ & nAP$_{75}$ & nAPs & nAPm & nAPl \\ \hline
        FRCN-ft-full~\cite{yan2019meta}$^{*}$    & 18.1 & 21.0 & 9.2  & 17.0  & 9.2   & 3.4  & 8.3  & 15.1 & 18.6 & 20.6 & 12.5 & 23.0  & 12.0  & 3.1  & 12.0 & 21.0 \\
        FRCN-BCE~\cite{yan2019meta}$^{*}$        & 29.2 & 36.8 & 6.4  & -     & -     & -    & -    & -    & 30.2 & 36.8 & 10.3 & -     & -     & -    & -    & -    \\
        TFA w/ fc~\cite{wang2020frustratingly}       & 27.9 & 33.9 & 10.0 & \subest{19.2}  & 9.2   & 3.9  & 8.4  & \subest{16.3} & 29.3 & 34.5 & 13.5 & 24.9  & 13.2  & 5.0  & 12.6 & 21.7 \\
        TFA w/ cos~\cite{wang2020frustratingly}      & 28.4 & 34.6 & 9.8  & 18.7  & 9.0   & \best{4.5}  & 8.8  & 15.8 & 29.9 & 35.3 & 13.6 & 25.0  & 13.4  & \best{5.9}  & 12.2 & 21.3 \\
        MPSR~\cite{wu2020multi}            & 15.3 & 17.1 & 9.7  & 17.9  & \subest{9.7}   & 3.3  & \subest{9.2}  & 16.1 & 17.1 & 18.1 & \subest{14.1} & \subest{25.4}  & \subest{14.2} & 4.0  & \subest{12.9} & \subest{23.0} \\
        Meta R-CNN~\cite{yan2019meta}     & 5.4  & 5.2  & 6.1  & 19.1  & 6.6   & 2.3  & 7.7  & 14.0 & 7.8  & 7.1  & 9.9  & 25.3  & 10.8  & 2.8  & 11.6 & 19.0 \\
        FsDetView~\cite{xiao2020few}      & 6.7  & 6.4  & 7.6  & -     & -     & -    & -    & -    & 10.0 & 9.3  & 12.0 & -     & -     & -    & -    & -    \\
        Retentive R-CNN~\cite{fan2021generalized} & \best{32.1} & \best{39.2} & \best{10.5} & \best{19.5} & 9.3 & 3.9 & 8.5 & \subest{16.3} & \subest{32.9} & \subest{39.3} & 13.8 & 22.9 & 13.8 & 4.6 & 11.9 & 22.6 \\ \hline
        \approach{} & \subest{32.0} & \best{39.2} & \subest{10.3} & 18.7 & \best{9.9} & \best{4.5} & \best{10.0} & \best{16.8} & \best{33.1} & \best{39.4} & \best{14.2} & \best{26.2} & \best{14.8} & \subest{5.3} & \best{13.1} & \best{23.9} \\
    \hlineB{3}
    \multicolumn{17}{l}{$^{*}$: results are reported by Retentive R-CNN~\cite{fan2021generalized} and TFA~\cite{wang2020frustratingly}.}
    \end{tabular}
    }\label{tab:coco}
\end{table*}

\subsection{Datasets \& Training Details}\label{sec:evalsetup}

\bitem{Pascal VOC}~\cite{pascal-voc-2007,pascal-voc-2012} consists of 20 classes where the class split for $\basecls{}$ and $\novelcls{}$ are 15 and 5 separately. 
The train set $\baseset{} \cup \novelset{}$ are from Pascal VOC 07+12 trainval sets~\cite{pascal-voc-2007,pascal-voc-2012} where $\novelset{}$ is randomly sampled with $K$ in $\{1,2,3,5,10\}$.
Following TFA~\cite{wang2020frustratingly}, we conduct experiments on three base-novel class partitions marked as $\{1,2,3\}$. In each partition, for fair comparison, we use the same sampled novel instances and report the detection precision for $\novelcls{}$ (nAP$_{50}$), $\basecls{}$ (bAP$_{50}$) and $\basecls{} \cup \novelcls{}$ (AP$_{50}$) on Pascal VOC 07 test set~\cite{pascal-voc-2007}. 

\bitem{MS COCO}~\cite{lin2014microsoft} is derived from COCO14~\cite{lin2014microsoft} consisting of 80 classes where $|\basecls{}|=60,|\novelcls{}|=20$ and $\novelcls{}$ are in common with Pascal VOC. The $\baseset{}$ and $\novelset{}$ are from train set with $K=\{10,30\}$. The detection precision of $\novelcls{}$ (nAP), $\basecls{}$(bAP) and $\basecls{} \cup \novelcls{}$ (AP) on COCO 14 val set are reported.

\bitem{LVIS}~\cite{gupta2019lvis} is derived from COCO17~\cite{lin2014microsoft} and contains $\sim$0.7M training instances of 1230 classes. 
The classes are divided into three groups w.r.t. the amount of annotation, rare (1-10), common (11-100), and frequent ($>$100).
Following~\cite{wang2020frustratingly}, we report the precision for all classes (AP) and class groups (AP$_r$, AP$_c$, and AP$_f$) on the val set.

\bitem{Implementation Details.} We instanlize our approach on Faster-RCNN~\cite{wang2020frustratingly,ren2015faster} which employs a region proposal network (RPN) to generate region candidates. For fair comparison, we use ResNet-101 with FPN~\cite{lin2017feature} as backbone to extract image feature maps where the Resnet-101 backbone is initialized by ImageNet~\cite{krizhevsky2017imagenet}-pretrained model.
As the outputs of penultimate layer in original classification module are non-negative and does not meet the property of the ETF classifier, we add a linear layer (projector) with the same input and output dimension on top of the penultimate layer. The projector output is then used for classification.
For RFS~\cite{he2009learning}, we set the up-sampling threshold as 0.01 for PASCAL VOC and MS COCO and 0.001 for LVIS.
During distillation, we share and fix the parameters of ResNet101 and FPN and only learn a new detection head.
We follow the setup in TFA~\cite{wang2020frustratingly} baseline such as SGD optimizer~\cite{sutskever2013importance}. More details can be found in Supp.

\subsection{Comparison with FSOD Methods}\label{sec:experiment}

We show the comparisons between our methods and state-of-the-art few-shot object detection approaches on PASCAL VOC and MSCOCO. We follow previous works to conduct experiments on three data splits with different shots of novel classes.
As for the performance AP$_{50}$ over all classes in Table~\ref{tab:voc-all}, our~\approach{} achieves the best performances for 12 out of 15 cases. Compared to the baseline method TFA~\cite{wang2020frustratingly}, our~\approach{} outperformed TFA consistently in all shots \& splits. Compared to the state-of-the-art Retentative RCNN model, our~\approach{} achieves better AP$_{50}$ when the number of shots is larger than 2, and obtains comparable performances for extremely few-shot cases. As for the detailed comparisons over novel classes (nAP$_{50}$) and base classes (bAP$_{50}$) in Table~\ref{tab:voc-all-detail}, our~\approach{} still consistently outperforms the baseline TFA method for all the settings. 

In addition, our~\approach{} achieves a better trade-off between base-class performance and novel-class generalization. On the one hand, although MPSR achieved higher performance for extremely few-shot settings (\eg, 36.2 vs. 31.6 for 1-shot) and competitive performances with $\{3,5,10\}$-shots, its performance drops by large margins (\eg, 68.1 vs. 81.3 for bAP$_{50}$). This observation indicates that MPSR improves the few-shot generalization with the sacrifice of base-class knowledge. On the other hand, as Retentive RCNN~\cite{fan2021generalized} includes the base detector \& RPN through max ensembling at test time, the adapted detector can be trained to specifically detect novel instances, where the bAP$_{50}$ is slightly higher than ours (\ie, 81.6 vs. 81.3). However, its few-shot generalization is not satisfying when the number of shots is larger than 2 (\eg, 6.3 lower than ours with 10 shots). In contrast, our approach only train a single detector and achieves stable and consistent gain. Similarly, for the results on MSCOCO shown in Table~\ref{tab:coco}, our~\approach{} outperforms Retentive RCNN for the novel-class metrics including nAP$_{75}$, nAPs, nAPm and nAPl. These comparison demonstrate that our~\approach{} has a strong few-shot generalization ability without base-class knowledge forgetting.

\begin{figure*}[]
    \centering
    \includegraphics[width=0.9\linewidth]{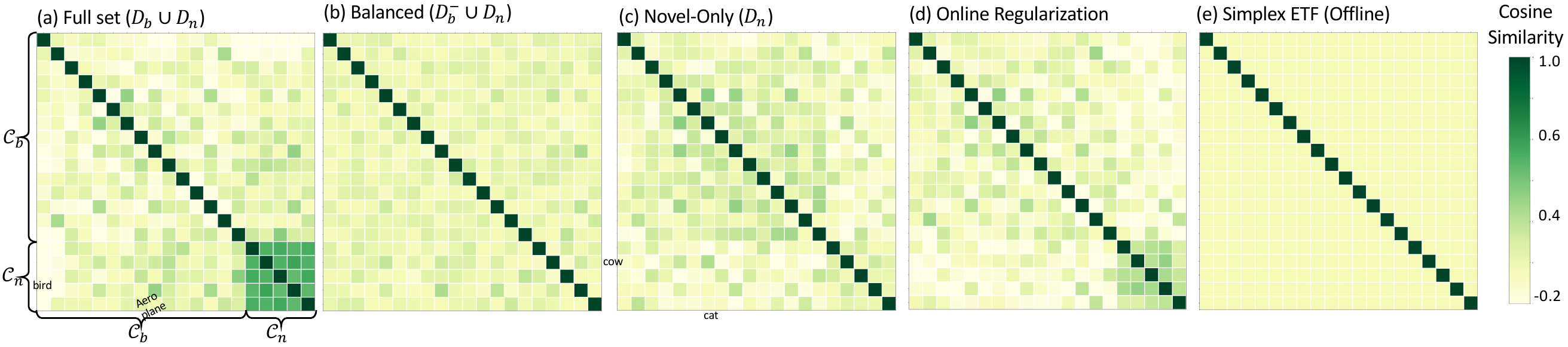}
    \caption{Comparison of pair-wise similarities of classifier weights. We follow the TFA~\cite{wang2020frustratingly} baseline but alter the train set at adaptation step between (a) full set $\baseset{} \cup \novelset{}$, (b) a balanced set $\basesubset{} \cup \novelset{}$, or (c) \textit{novel} set $\novelset{}$. For our approach, to maximally separate the classifier weights, we can (d) apply Eq.~\eqref{eq:etf} as a regularization term during training (online) or (e) derive the classifier weights offline and fix them. The weights for all classes are arranged w.r.t a decreasing order of the number of annotations per class, \ie, from top (left) to bottom (right), from many-shot to few-shot.}
    \label{fig:step-center-ablation}
\end{figure*}

\subsection{Analysis of Inter-Class Separation}\label{sec:strategy-generalization}

\noindent \textbf{Revisit the conventional adaptation strategy from the perspective of separation between classes.}
Recall that existing few-shot object detection methods follow in TFA~\cite{wang2020frustratingly} and employs a two-step strategy, \ie, first pre-train on the base train set $\baseset{}$ to learn a \textit{Base} detector, and then fine-tune on the union of the downsampled base set and the novel set, \ie, $\basesubset{} \cup \novelset{}$. We take TFA~\cite{wang2020frustratingly} as the baseline and consider the following settings for the second step: (1) full set $\baseset{} \cup \novelset{}$, (2) balanced set $\basesubset{} \cup \novelset{}$, (3) only novel set  $\novelset{}$.

As shown in Fig.~\ref{fig:step-center-ablation}, we visualize the separation of classifier weights based on their pair-wise cosine similarities.
By comparing Fig.~\ref{fig:step-center-ablation}(a-c), fine-tuning among a balanced set is vital to learn the well-separated classifier weights for all classes $\basecls{} \cup \novelcls{}$.
Instead, using the full set would make the novel classes entangled in Fig.~\ref{fig:step-center-ablation}(a) due to the extremely imbalanced class distribution (\ie, $|\novelset{}| \ll |\baseset{}|$). Although only using $\novelset{}$ can maximally separate the weights of $W_n$, as no training data of $\basecls{}$ is seen, the separation between \textit{base} weights in $W_b$ is hurt and each \textit{novel} class center $\mathbf{w}_i \in W_n$ may still close to some \textit{base} weight, \eg, the similarity between classes ``cat'' and ``cow'' is relatively high in Fig.~\ref{fig:step-center-ablation}(c).

As summarized in Table~\ref{tab:step}, the novel detection fails when the full set is used in fine-tuning (\tablerow{1}). The detection precision on both $\basecls{} \cup \novelcls{}$ is sub-optimal when no annotation of base class is provided (\tablerow{3}) and the detector can easily overfit to the small $\novelset{}$. Then, finetuning on the balanced set (\tablerow{2}) can preserve the base knowledge, maximize the few-shot adaptation effect, and achieve the highest score among the three settings. However, such a balanced set has discarded the diverse training samples of $\basecls{}$ and the performance drop in base detection is inevitable.

\noindent \textbf{Training on the union of whole base set and upsampled novel set.}
In contrast, we propose to train from $\baseset{} \cup \novelsubset{}$ directly. Note that $\novelsubset{}$ is a duplication of $\novelset{}$ with same images but more copies. To properly separate the features of different classes, we use the data-independent optimization target in Eq.~\ref{eq:etf} to derive a ETF classifier weights $W^*$ offline. 
As mentioned in Sec.~\ref{sec:inter}, Eq.~\ref{eq:etf} can still be used as a regularization loss to supervised the learning of the last linear layer during training (online). However, as shown in Fig.~\ref{fig:step-center-ablation}(d), it is still hard to get a perfect ETF classifier shown in Fig.~\ref{fig:step-center-ablation}(e). After all, the update of classifier weights is also impacted by the weight decay regularization and classification loss, and the learning of weights is not stable, in particular, on an extremely imbalanced dataset. 
As the classifier weights are kept being updated, the optimization direction of each feature cluster is not stable, which then impede the adaptation efficiency. As compared in Table.~\ref{tab:online}, the performance by online optimization is slightly worse, in particular when $K=1$.
Though the classifier weights are fixed in ETF, as the pair-wise angles between weights are the same, we can equivalently assign the weights to all classes $\mathcal{C}$.

\begin{table}[]
\centering
    \caption{Comparison between adaptation strategies.}
    \resizebox{\linewidth}{!}
    {\renewcommand{\arraystretch}{1.03}
    \begin{tabular}{l|cccccc}
    \hlineB{3}
        \multirow{2}{*}{Approach} & \multicolumn{3}{c}{VOC 5-shot} & \multicolumn{3}{c}{COCO 10-shot} \\
                                  &  AP$_{50}$   & bAP$_{50}$  & nAP$_{50}$  & AP   & bAP  & nAP  \\ \hline
        $\baseset{}\cup \novelset{}$ & 60.6 & 80.8 & 0 & 29.4 & 39.2 & 0 \\
        $\novelset{}$ & 43.6 & 74.8 & 44.1 & 24.1 & 29.5 & 7.8 \\
        $\basesubset{}\cup \novelset{}$ & 73.4 & 79.3 & 55.7 & 27.9 & 33.9 & 10 \\ \hline
        \fixstage{} & 74.9 & 81.0 & 56.4 & 31.5 & 38.8 & 9.6 \\
        \approach{} & \textbf{75.4} & \textbf{81.0} & \textbf{58.7} & \textbf{32.0} & \textbf{39.2} & \textbf{10.3} \\
    \hlineB{3}
    \end{tabular}
    }\label{tab:step}
\end{table}
\begin{table}[]
\centering
    \caption{Comparison between Offline and Online Classifiers}
    \resizebox{\linewidth}{!}
    {\renewcommand{\arraystretch}{1.0}
    \begin{tabular}{l|ccc|ccc}
    \hlineB{3}
        \multirow{2}{*}{Approach} & \multicolumn{3}{c|}{1-shot} & \multicolumn{3}{c}{5-shot} \\
         & AP$_{50}$ & bAP$_{50}$ & nAP$_{50}$ & AP$_{50}$ & bAP$_{50}$ & nAP$_{50}$ \\ \hline
        Offline & \textbf{68.9} & 79.9 & \textbf{35.8} & \textbf{74.9} & \textbf{81.0} & \textbf{56.4} \\
        Online & 68.4 & \textbf{80.2} & 33.1 & 74.6 & 80.8 & 55.9 \\
    \hlineB{3}
    \end{tabular}
    }\label{tab:online}
\end{table}

Next, as compared in Table~\ref{tab:ablation} \tablerow{1,3,4}, though adding margins or performing RFS may help with inter-class separation and improve nAP$_{50}$ on $\novelcls{}$, since the weights $W_n$ are still not well-learned due to the extreme imbalance between $\baseset{}$ and $\novelset{}$, the performance gain is limited. In contrast, fixing the weights as ETF (\tablerow{(3,5),(4,6)}) can improve the novel detection, in particular, the nAP$_{50}$ is boosted from 12.2 to 35.8 in \tablerow{3,5}, which shows that the inter-class separation is essential for distinguishing objects in GFSOD.

\begin{table}[]
\centering
    \caption{Ablation study of the \fixstage{} baseline.}
    \resizebox{0.95\linewidth}{!}
    {\renewcommand{\arraystretch}{1.0}
    \begin{tabular}{c|ccc|ccc}
    \hlineB{3}
        Idx. & ETF & Margin & RFS  & AP$_{50}$   & bAP$_{50}$  & nAP$_{50}$ \\ \hline
        1&  & & & 60.6 & 80.8 & 0.0\\
        2& \checkmark{} & & & 61.8 & 81.2 & 3.6\\
        3&  & \checkmark{} & & 63.9 & 81.1 & 12.2\\
        4&  & & \checkmark{} & 62.8 & 80.8 & 9.3\\
        5& \checkmark{} & \checkmark{} & & 69.6 & 81.0 & 35.8\\
        6& \checkmark{} & & \checkmark{} & 65.2 & \textbf{81.2} & 17.0\\
        7&\checkmark{} & \checkmark{} & \checkmark{} & \textbf{74.9} & 81.0 & \textbf{56.4}\\
    \hlineB{3}
    \end{tabular}
    }\label{tab:ablation}
\end{table}

\begin{table}[!ht]
    \centering
    \caption{Comparison of Initialization.}
    \resizebox{\linewidth}{!}
    {
        \begin{tabular}{ll|cccccc}
        \hlineB{3}
            \multirow{2}{*}{Init.} & \multirow{2}{*}{Method} & \multicolumn{3}{c}{VOC 5-shot} & \multicolumn{3}{c}{COCO 10-shot} \\
              &  & AP$_{50}$   & bAP$_{50}$  & nAP$_{50}$  & AP   & bAP  & nAP  \\ \hline
            Base & \fixstage{}  & 74.9 & 80.9 & 56.7 & 31.7 & 39.0 & 9.7\\
            Base & DeFRCN~\cite{qiao2021defrcn} & 74.1 & 77.1 & 65.1 & 30.1 & 34.4 & 17.3\\
            \fixstage{} & DeFRCN~\cite{qiao2021defrcn}  & 74.8 & 78.0 & 65.3 & 30.7 & 35.1 & 17.4 \\
        \hlineB{3}
        \end{tabular}
    }\label{tab:initialization}
\end{table}

Furthermore, being orthogonal to the previous FSOD approaches, our model can be intuitively used as initialization for their adaptation. For the sake of simplicity, we only consider \fixstage{} and use a strong baseline DeFRCN~\cite{qiao2021defrcn} for comparison. The PCB calibration~\cite{qiao2021defrcn} is removed to better demonstrate the effect of \fixstage{}. As reported in Table~\ref{tab:initialization}, though DeFRCN has improved \textit{novel} detection (nAP$_{50}$) significantly, it still sacrifices the performance on base set. Then, comparing with using \textit{Base} detector as initialization, on both datasets, using our \fixstage{} can both help with the adaptation on $\novelcls{}$ and mitigate the drop in $\basecls{}$ (bAP$_{50}$).
Finally, comparing Table~\ref{tab:initialization} \tablerow{1} and \fixstage{} in Table~~\ref{tab:ablation}, adding the step of \textit{Base} detector initialization can only provide marginal improvement for \fixstage{}. As \fixstage{} has already outperformed TFA, we skip pre-training step for simplicity.

\subsection{Analysis of Intra-Class Compactness}\label{sec:discussion}

\begin{figure}[t]
    \centering
    \includegraphics[width=0.99\linewidth]{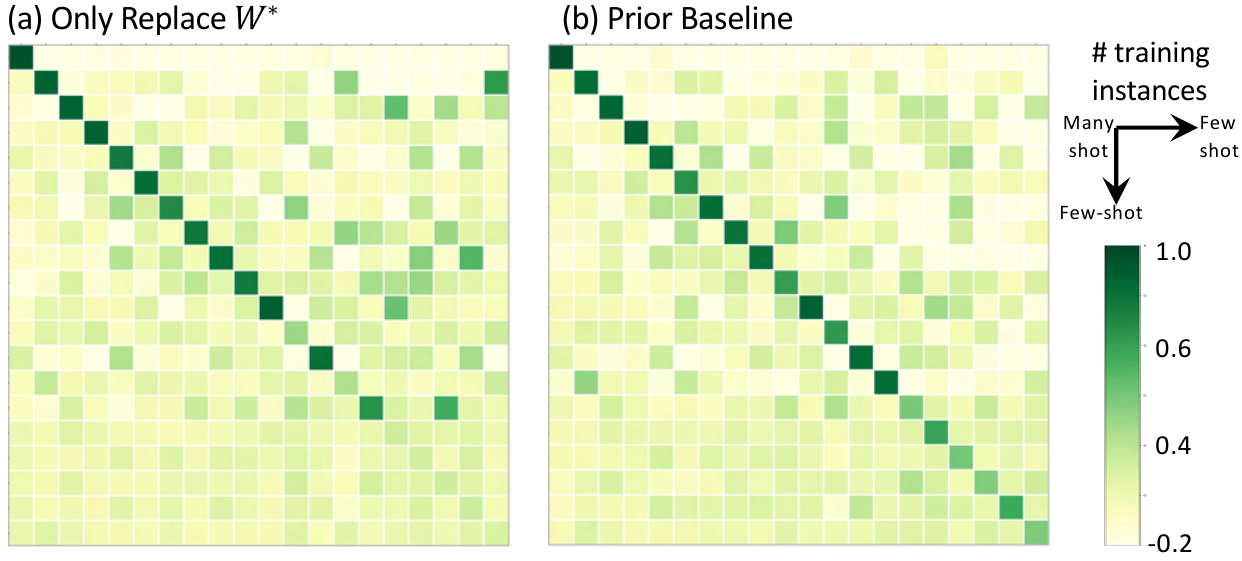}
    \caption{Pair-wise cosine similarity between class means and weights. The class mean is obtained by averaging of features for each class. When the detector is trained on full set $\baseset{} \cup \novelset{}$, we first (a) replace the linear classifier with ETF and (b) then use our \fixstage{} baseline. Our \fixstage{} can push the features close to the assigned weights effectively.}
    \label{fig:step-center-mean}
\end{figure}

Even though the classifier weights have been maximally and equally separated in Fig.~\ref{fig:step-center-ablation}(e), as the training data is limited, it is still necessary to effectively push the features towards the assigned weight. As compared in Fig.~\ref{fig:step-center-mean}, when ETF is used, for each $c \in \novelcls{}$, as $|\novelset{}| \ll |\baseset{}|$, the mean of the its features is still distant from the assigned weights. However, our \fixstage{} baseline clearly push the features to the assigned weights to facilitate the novel detection.
Similarly, in Table~\ref{tab:ablation}, only using the ETF classifier can introduce limited gain (\tablerow{1,2}). Though the ETF classifier with dot-regression loss has been used for long-tail classification~\cite{yang2023neural,yang2022we}, we note the efficiency in dealing with hugely imbalanced datasets is limited. By adding margins to tighten each cluster and/or up-sampling novel instances in RFS to ensure that sufficient features of $\novelcls{}$ are used for training, the nAP$_{50}$ can then be improved (\tablerow{2,5,6,7}).

\begin{figure}
    \centering
    \includegraphics[width=\linewidth]{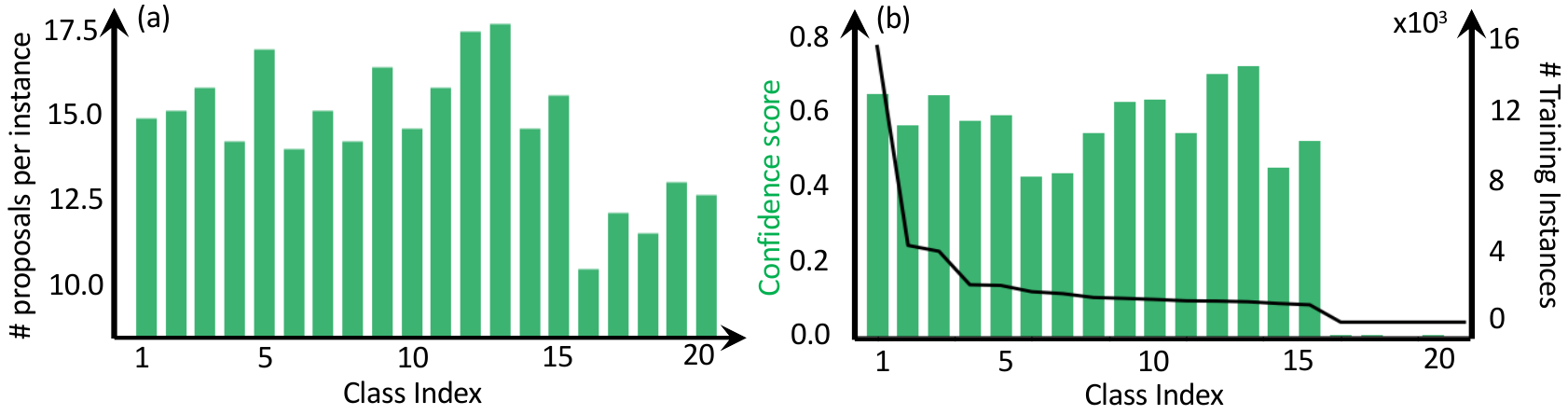}
    \caption{With a detector trained on the $\baseset{} \cup \novelset{}$ with vanilla $\mathcal{L}_{CE}$. For each class, we calculate the mean of a) the number of proposals per instance and b) classification confidence score where the number of annotations per class is plotted for reference.}
    \label{fig:stat}
\end{figure}

\bitem{Obtaining effective margins} is essential to train on an extremely imbalanced dataset. As discussed in \cite{menon2020long}, the margins to be added should meet conditions such as \textit{Fisher consistency}~\cite{lin2004note,bartlett2006convexity} to balance the error among different classes.
As $|\novelset{}| \ll |\baseset{}|$, directly learning the margins for each class individually from scratch (Table~\ref{tab:margin} \tablerow{1}) is difficult and may suffer from training instability such as gradient explosion. By sharing margins for classes in the same group, \ie, $\basecls{}$, $\novelcls{}$, and $\negcls{}$, the nAP$_{50}$ can be improved slightly.

As summarized in Fig.~\ref{fig:stat}, for each class, the number of proposals used to train the detection module ranges from 11 to 17 per instance on average. As such, using the prior of instance distribution $\{p_c\}_{c \in \basecls{} \cup \novelcls{}}$ can help estimate good margins (\fixstage{}). However, as the number of proposals for $\novelcls{}$ (11$\sim$14) is still slightly less than that of $\basecls{}$ (13$\sim$17) and the margin $m_{\negcls{}}$ for $\negcls{}$ is roughly estimated, it is still necessary to learn margins adaptively. As no stronger prior knowledge can be used, directly learning the margins initialized by $\{-\log(p_c)\}$ does not help clearly (\fixstage{}+). However, through self-distillation, the logits output by pretrained \fixstage{} baseline model can be used to indicate the relationship between the proposals features and all class centers, which is then used as supervision signal in our \approach{}.

\subsection{Extension to Long-tailed Object Detection}

\begin{table}[]
\centering
    \caption{Ablation study of learning effective margins.}
    \resizebox{\linewidth}{!}
    {\renewcommand{\arraystretch}{1.0}
    \begin{tabular}{ccc|ccc}
    \hlineB{3}
        Approach & Fixed$^*$ & Init. & AP$_{50}$   & bAP$_{50}$  & nAP$_{50}$ \\ \hline
        individual & & Random & 56.4 & 69.5 & 17.1\\
        group & & Random &  60.4 & 73.2 & 21.9\\
        \fixstage{} & \checkmark{} & $-\log(p_c)$ & 74.9 & \textbf{81.0} & 56.4\\
        \fixstage{}+ & & $-\log(p_c)$ & \textbf{74.8} & 80.8 & \textbf{56.7}\\
    \hlineB{3}
    \multicolumn{6}{l}{$^*$: The margins are fixed during training.}
    \end{tabular}
    }\label{tab:margin}
\end{table}

\begin{table}[]
\centering
    \caption{Performance comparison of LVIS.}
    \resizebox{\linewidth}{!}
    {\renewcommand{\arraystretch}{1}
    \begin{tabular}{l|cccc|c}
    \hlineB{3}
        \multirow{2}{*}{Approach} & \multicolumn{4}{c|}{ResNet-50} & ResNet-101 \\
                    & AP    & APr   & APc   & Apf   & AP   \\ \hline
        Baseline~\cite{gupta2019lvis} & 22.7  & 10.6  & 22.0  & 28.0  & 24.5   \\
        TFA w/ fc~\cite{wang2020frustratingly} & 24.1  & 14.9  & 23.9  & 27.9  & -    \\
        TFA w/ cos~\cite{wang2020frustratingly}  & 24.4  & 16.9  & 24.3  & 27.7  & -    \\
        \approach{} (Ours) & \textbf{24.9}  & \textbf{17.3}  & \textbf{24.6}  & \textbf{28.5}  & \textbf{26.8} \\ \hline
        RFS~\cite{gupta2019lvis}         & 24.9  & 14.4  & 24.5  & 29.5  & -    \\
        Focal Loss~\cite{lin2017focal}  & 22.0  & 10.5  & 22.4  & 25.9  & -    \\
        EQL~\cite{tan2020equalization}         & 25.1  & 11.9  & 26.0  & 29.1  & 26.1 \\
        BAGS~\cite{li2020overcoming}        & 26.0  & 17.7  & 25.8  & \textbf{29.5}  & 26.4 \\
        ACSL~\cite{wang2021adaptive}        & 26.4  & 18.6  & 26.4  & 29.4  & 27.5 \\
        \approach{} (Ours) & \textbf{26.7}  & \textbf{18.9}  & \textbf{27.0}  & 29.0  & \textbf{27.9} \\ \hline
    \hlineB{3}
    \multicolumn{6}{l}{Full table can be found in the Supp.}
    \end{tabular}
    }\label{tab:lvis}
\end{table}

As compared in Table~\ref{tab:lvis}, we use TFA~\cite{wang2020frustratingly} and ACSL~\cite{wang2021adaptive} as two baselines. By employing our design, our \approach{} can achieve higher detection precision on both cases. For comparison with ACSL, we follow the training procedure in ACSL and our approach can benefit from the prior of data distribution to learn discriminative features.
More detailed explanation \& results can be found in Supp.

\section{Conclusion}
In this paper, we revisit generalized few-shot object detection from a perspective of discriminative feature learning. We further proposed a simple but effective framework, Discriminative Geometry-aware (\approach{}) learning, for inter-class separation and intra-class compactness. Experiments demonstrates that our~\approach{} improves generalization on novel classes without hurting the detection of base classes, and can be extended to long-tail object detection. In the future, we will keep investigating the desired properties of features in object detection and adapted it more realistic scenarios such as domain adaptation.

{\small
\begin{spacing}{0.9}
\noindent \textbf{Acknowledgement} This material is based on research sponsored by Air Force Research Laboratory (AFRL) under agreement number FA8750-19-1-1000. 
The U.S. Government is authorized to reproduce and distribute reprints for Government purposes notwithstanding any copyright notation therein. 
The views and conclusions contained herein are those of the authors and should not be interpreted as necessarily representing the official policies or endorsements, either expressed or implied, of Air Force Laboratory, DARPA or the U.S. Government.
\end{spacing}
}

\section{Approach}

\subsection{Simplex ETF \& Neural Collapse}

The neural collapse (NC) phenomenon is revealed by \cite{papyan2020prevalence} in the fully-supervised learning, \ie, an extremely simple mathematical structure on the last-layer features. 
In particular, when the model is well-trained on a balanced dataset, the $N_f$ features $\{\mathbf{x}_{c,i}\}_{i=1}^{N_f}$ for class $c$ will converge to its class mean $\bar{\mathbf{x}}_c = \frac{1}{N_f} \sum_{i} \mathbf{x}_{c,i}$ where the class means $\{\bar{\mathbf{x}}_c\}_{c \in \mathcal{C}}$ together the class centers $\{\mathbf{w}_c\}_{c \in \mathcal{C}}$ will collapse to the simplex equiangular tight frame (Simplex ETF). Meanwhile, though the optimization objective of class mean and class centers (classifier weights) are not exactly the same, the class mean and class centers will still converge to each other.

\noindent \textbf{\emph{Simplex Equiangular Tight Frame}} denotes a collection of vectors $\mathbf{W}^* = \{\mathbf{w}_i'\}_{i=1}^{N_c} \in \mathcal{R}^{d \times N_c}$ that 
\begin{equation}\label{eq:definition}
    \mathbf{W}^* = \sqrt{\frac{{N_C}}{{N_C}-1}} \mathbf{U} \big( \mathbf{I}_{N_C} - \frac{1}{N_c} \mathbf{1}_{N_C} \mathbf{1}_{N_C}^T \big)
\end{equation}
where each vector $\mathbf{w}_i' \in \mathcal{R}^d$ and $\|\mathbf{w}_i'\|_2 = 1$ for $1 \leq i \leq N_C$, $\mathbf{I}_{N_C} \in \mathcal{R}^{N_C \times N_C}$ and $\mathbf{1}_{N_C} \in \mathcal{R}^{N_C}$ denote the identity matrix and all-ones vector respectively. The rotation matrix $\mathbf{U} \in \mathcal{R}^{d \times N_C}$ satisfies $\mathbf{U}^T \mathbf{U} = \mathbf{I}_{N_C}$ and $d \geq N_C-1$.
In this way, for all vectors in a simplex ETF, their pair-wise angles are identical, \ie,
\begin{equation}\label{eq:equiangular}
    \mathbf{w}_i'^T \mathbf{w}_j' = -\frac{1}{{N_C}-1}, \forall i,j \in [1,{N_C}]~~\text{and}~~i \neq j,
\end{equation}
where the angle $arc cos(-\frac{1}{{N_C}-1})$ shown in \cite{papyan2020prevalence} ans is the maximal equiangular angle of $N_C$ vectors in the feature space. 

\begin{algorithm*}[h]
    \caption{Iterative algorithm for obtaining Simplex ETF)}
    \begin{algorithmic}\label{Alg:ETF-iterative}
        \STATE \textbf{Input:} Number of classes $N_C$, feat. dim $d$ where $d \leq N_C-1$  maximum iterations $T$, stop threshold $\delta$, learning rate $\tau$.
        \STATE 1:~~Initialization: 
        \STATE ~~~~~~~~~Randomly initialize $W$ = np.random.normal(size=($N_C,d$)), 
        \STATE 2:~~for $t=1$ to $T$
        \STATE ~~~~~~~~~$l_2$-normalize the vector $\mathbf{w}$ in each row $W$ = normalize($W$),
        \STATE ~~~~~~~~~calculate pair-wise $l_2$ distance $W_{inner} \in \mathcal{R}^{N_C \times N_C}$ and set diagonal value as max infinity
        \STATE ~~~~~~~~~calculate minimum distance for each class (except for the diagonal value) $idx = np.argmin(W_{inner}, axis=1)$
        \STATE ~~~~~~~~~calculate objective funcdtion $obj = np.sum(W_{inner}[range(N_C), idx])$
        \STATE ~~~~~~~~~calculate gradient $grad = (W - W[nn_index,:]) * 2$ 
        \STATE ~~~~~~~~~update the weight $W' = W + grad * \tau$
        \STATE ~~~~~~~~~$l_2$-normalize the vector $\mathbf{w}'$ in each row $W'$ = normalize($W'$),
        \STATE ~~~~~~~~~update the weight $W' = W$,
        \STATE ~~~~~~~~~if $obj < \tau $:
        \STATE ~~~~~~~~~~~~~~~~~~Early stop, set T as t
        \STATE ~~~~~end for
        \STATE \textbf{Output: } $\mathcal{X}^{(T+1)}$ 
    \end{algorithmic}
\end{algorithm*}

Note that the equation in Eq.~\ref{eq:definition} is a \emph{closed-form} for obtaining an ETF but it is only used when $d \leq N_C$. When $d = N_C+1$, we can then use an \emph{iterative algorithm} to obtain the ETF. Specifically, we randomly initialize the values in $W$ and use the Eq.1 in the main paper to update the weight values. We provide the python-stype pseudo-code below. Note, since we want to maximize the objective function, we apply $+$ in the weight updating part.

\section{Experiment}

\subsection{Implementation Details}

The Faster-RCNN system we are using consists of a ResNet-101 feature backbone, a RPN network, and a detection module. The detection module is used to extract features for each region proposal, a linear classifier and a regression for localization. As mentioned in the main paper, 
since penultimate layer in the classification module is followed by a ReLU activation~\cite{goodfellow2017deep}, the proposal features are constrained to have non-negative entries and its distance to weights in $W^{*}$ are lower-bounded, and we thus add a linear layer (projector) on top of the extractor of proposal feature. Meanwhile, as highlighted in Sec. 5.3 in the main paper, we do not need to pretrained the detector on the base set, but directly training everything from scratch, however, we will still use the ImageNet-pretrained model to initialize the feature extractor.

The dimension of proposal feature in Faster-RCNN is $d=1024$ by default. As such, for experiments on MSCOCO and Pascao VOC, we set the projector with the same input and output dimension. However, for experiments on LVIS, since it has 1230 classes in v0.5 and 1203 classes in v1.0, we set the output dimension of projector as 1280.

During distillation, as we mainly focus on the learning of detector. As such, we fix the ResNet-101 feature backbone and the RPN network, and only distill the detection module. Also, during distillation, we do not apply any distillation strategy on the layer for localization. Then, we first use the fixed margin $-\log (p_c)$ in the loss and train the whole network. Then, during distillation, to fasten the training process, we can choose to also initialize the detector module with the pretrained teacher model. 

\paragraph{RFS implementation details.} We directly call the ``RepeatFactorTrainingSample'' as the training sampler function and send rfs parameter (0.01 for VOC \& COOC and 0.001 for LVIS) to the variable ``SAMPLER TRAIN''

\subsection{Full experiment on Pascao VOC}

We summarize the performance of novel detection in Table~\ref{tab:voc-novel}. 
Comparing with baseline TFA, over all 15 experiments on PASCAL VOC, the \fixstage{} baseline has already outperformed TFA by 3.6 gain in nAP$_{50}$ and 1.9 gain in bAP$_{50}$ on average. By performing the self-distillation to adjust margins for all classes $\mathcal{C}$ adaptively, our full approach \approach{} can further improve the detection score, \eg, comparable nAP$_{50}$ with MPSR~\cite{wu2020multi} but maintaining high \textit{base} detection precision (81.3 Vs. 68.1).
As reported in Table 2 in the main paper, comparing with Retentive RCNN~\cite{fan2021generalized}, a state-of-the-art (SOTA) approach in GFSOD, besides maintaining precise base detection, our approach also improves the novel detection score (43.9 Vs. 41.1). Meanwhile, the superior performance by Retentive RCNN on split 1 when $K$ is $\{1,2\}$ cannot be generalized to other splits. However, our approach achieves stable and consistent gain. Meanwhile, when more training data are provided, \ie, $K \geq 3$, the advantage of our \approach{} is better explored and achieve 3.76 nAP$_{50}$ gain on average.

\begin{table*}[h]
\centering
    \renewcommand\thetable{M1}
    \caption{Performance comparison of nAP$_{50}$ on the PASCAL VOC dataset.}
    \resizebox{\linewidth}{!}
    {\renewcommand{\arraystretch}{1.1}
    \begin{tabular}{l|ccccc|ccccc|ccccc|c}
    \hlineB{3}
        \multirow{2}{*}{Approach} & \multicolumn{5}{c|}{split 1}      & \multicolumn{5}{c|}{split 2}      & \multicolumn{5}{c|}{split 3}  & \multirow{2}{*}{Avg.} \\ 
                                   & 1    & 2    & 3    & 5    & 10   & 1    & 2    & 3    & 5    & 10   & 1    & 2    & 3    & 5    & 10 &   \\ \hline
        FRCN-ft-full~\cite{yan2019meta}$^{*}$    & 15.2 & 20.3 & 29   & 25.5 & 28.7 & 13.4 & 20.6 & 28.6 & 32.4 & 38.8 & 19.6 & 20.8 & 28.7 & 42.2 & 42.1 & 27.1\\
        TFA w/ fc~\cite{wang2020frustratingly}   & 36.8 & 29.1 & 43.6 & 55.7 & 57   & 18.2 & 29   & 33.4 & 35.5 & 39.0   & 27.7 & 33.6 & 42.5 & 48.7 & 50.2 & 38.7 \\
        TFA w/ cos~\cite{wang2020frustratingly}  & 39.8 & 36.1 & 44.7 & 55.7 & 56   & 23.5 & 26.9 & 34.1 & 35.1 & 39.1 & 30.8 & 34.8 & 42.8 & 49.5 & 49.8 & 39.9\\
        MPSR~\cite{wu2020multi}                  & \best{42.8} & \subest{43.6} & \subest{48.4} & 55.3 & \subest{61.2} & \best{29.8} & 28.1 & \subest{41.6} & \best{43.2} & 47.0 & \best{35.9} & \subest{40.0} & 43.7 & 48.9 & 51.3 & \best{44.0} \\
        Meta RCNN~\cite{yan2019meta}$^{*}$       & 16.8 & 20.1 & 20.3 & 38.2 & 43.7 & 7.7  & 12.0 & 14.9 & 21.9 & 31.1 & 9.2  & 13.9 & 26.2 & 29.2 & 36.2 & 22.8 \\
        FSRW~\cite{perez2020incremental}         & 14.8 & 15.5 & 26.7 & 33.9 & 47.2 & 15.7 & 15.3 & 22.7 & 30.1 & 39.2 & 19.2 & 21.7 & 25.7 & 40.6 & 41.3 & 27.3 \\
        FsDetView~\cite{xiao2020few}$^{*}$       & 25.4 & 20.4 & 37.4 & 36.1 & 42.3 & 22.9 & 21.7 & 22.6 & 25.6 & 29.2 & \subest{32.4} & 19.0 & 29.8 & 33.2 & 39.8 & 29.2 \\
        Retentive R-CNN~\cite{fan2021generalized}& \subest{42.4} & \best{45.8} & 45.9 & 53.7 & 56.1 & 21.7 & 27.8 & 35.2 & 37.0 & 40.3 & 30.2 & 37.6 & 43.0 & 49.7 & 50.1 & 41.1\\ \hline
        \fixstage{} & 35.8 & 37.9 & 45.7 & \subest{56.4} & 61.0 & 22.7 & \subest{28.4} & 39.8 & 41.4 & \subest{48.8} & 30.8 & 36.4 & \subest{45.4} & \subest{52.4} & \subest{53.8} & 42.4\\
        \approach{} & 37.9 & 39.4 & \best{48.5} & \best{58.6} & \best{61.5} & \subest{26.6} & \best{28.9} & \best{41.9} & \subest{42.1} & \best{49.1} & 30.4 & \best{40.1} & \best{46.9} & \best{52.7} & \best{54.7} & \best{44.0}\\
    \hlineB{3}
    \multicolumn{17}{l}{$^{*}$: results reported by Retentive R-CNN~\cite{fan2021generalized} and TFA~\cite{wang2020frustratingly}.$^{\dagger}$: Model ensembling. Full tables can be found in Supp.}\\
    \end{tabular}
    }\label{tab:voc-novel}
\end{table*}

\begin{table*}[]
\centering
    \caption{Detailed Performance of MS COCO dataset.}
    \resizebox{\linewidth}{!}
    {\renewcommand{\arraystretch}{1.12}
    \begin{tabular}{l|cccccccc|cccccccc}
    \hlineB{3}
        \multirow{2}{*}{Approach} & \multicolumn{8}{c|}{10-shot} & \multicolumn{8}{c}{30-shot} \\
         & AP   & bAP  & nAP  & nAP$_{50}$ & nAP$_{75}$ & nAPs & nAPm & nAPl & AP   & bAP  & nAP  & nAP$_{50}$ & nAP$_{75}$ & nAPs & nAPm & nAPl \\ \hline
        \fixstage{} & 31.5 & 38.8 & 9.6  & 17.8  & 9.2   & 3.8  & 9.3  & 16.5 & 32.5 & 38.8 & 13.6 & 24.3 & 13.3 & 4.7 & 11.9 & 21.4 \\
        \approach{} & 32.0 & 39.2 & 10.3 & 18.7  & 9.9   & 4.5  & 10.0 & 16.8 & 33.1 & 39.4 & 14.2 & 26.2 & 14.8 & 5.3 & 13.1 & 23.9 \\
    \hlineB{3}
    \end{tabular}
    }\label{tab:coco-detail}
\end{table*}

\subsection{Long-Tail Object Detection}

\bitem{LVIS}~\cite{gupta2019lvis} is derived from COCO17~\cite{lin2014microsoft} and has two versions of annotations. The version v1.0 contains $\sim$1.3M training instances of 1203 classes while the version v0.5 has $\sim$0.7M training instances of 1230 classes. The one reported in the main paper is of v0.5. According to the number of training instances, the classes are divided into three groups, rare (1-10), common (11-100), and frequent ($>$100).
Following~\cite{wang2020frustratingly}, apart from the precision for all classes (AP) on the validation set, we also report the precision for each group, \ie, AP$_r$, AP$_c$, and AP$_f$.
Meanwhile, following a common setup, we try two different backbones ResNet50 and ResNet101.

Here we try two different baseline, TFA and ACSL. We do acknowledge other related research on LVIS such as EFL~\cite{li2022equalized} and LOCE~\cite{feng2021exploring}. However, these approaches are developed on Mask-RCNN framework, \ie, both object detection and object segmentation are trained. Since object segmentation introduces extra supervision signals, while our focus is main on object detection, we thus choose ACSL as the baseline. 

Comparing with ACSL, TFA also focus on object detection only but ACSL 1) applies a two step training strategy and 2) use the model pretrained on MSCOCO as initialization. In contrast, TFA only uses ImageNet-pretrained model to initialize the feature extractor. Meanwhile, it follow the configuration regarding learning rate and training epochs in the 1x Baseline but apply it on base training stage. As such, we consider both of these two setups. As such, we follow the training steps ACSL and use model pre-trained on MS COCO as initialization. From the Table ~\ref{tab:lvis-full}, \approach{} can achieve consistent gain on two cases.

\begin{table}[!ht]
\centering
    \renewcommand\thetable{M2}
    \caption{Performance comparison of LVIS dataset (Full Table)}
    \resizebox{\linewidth}{!}
    {\renewcommand{\arraystretch}{1.0}
    \begin{tabular}{l|cccccccc}
    \hlineB{3}
        \multirow{2}{*}{Approach} & \multicolumn{4}{c}{ResNet-50} & \multicolumn{4}{c}{ResNet-101} \\
                    & AP    & APr   & APc   & Apf   & AP     & APr   & APc   & Apf   \\ \hline
        \multicolumn{9}{c}{V0.5}                                                     \\ \hline
        1x Baseline & 22.7  & 10.6  & 22.0  & 28.0  & 24.5   & 13.1  & 23.9   & 30.0 \\ 
        TFA w/ fc~\cite{wang2020frustratingly} & 24.1  & 14.9  & 23.9  & 27.9  & -      & -     & -     & -     \\
        TFA w/ cos~\cite{wang2020frustratingly}  & 24.4  & 16.9  & 24.3  & 27.7  & -      & -     & -     & -     \\
        \approach{}        & 24.9  & 17.3  & 24.6  & 28.5  & 26.8   & 18.5  & 26.8  & 30.1  \\
        RFS~\cite{gupta2019lvis}         & 24.9  & 14.4  & 24.5  & 29.5  & -      & -     & -     & -     \\
        Focal Loss~\cite{lin2017focal}  & 22.0  & 10.5  & 22.4  & 25.9  & -      & -     & -     & -     \\
        EQL~\cite{tan2020equalization}         & 25.1  & 11.9  & 26.0  & 29.1  & 26.1   & 11.5  & 27.1  & 30.5  \\
        BAGS~\cite{li2020overcoming}        & 26.0  & 17.7  & 25.8  & 29.5  & 26.4   & 16.8  & 25.8  & 30.9  \\
        ACSL~\cite{wang2021adaptive}        & 26.4  & 18.6  & 26.4  & 29.4  & 27.5   & 19.3  & 27.6  & 30.7  \\
        \approach{} & 26.7  & 18.9  & 27.0  & 29.0  & 27.9   & 19.5  & 28.0  & 31.0  \\ \hline 
        \multicolumn{9}{c}{V1.0}                                                     \\ \hline
        1x Baseline & 19.3  & 6.4  & 17.1   & 27.6  & 21.1   & 10.1  & 21.7  & 25.8  \\
        \approach{} & 22.5  & 12.4 & 20.6   & 26.8  & 24.4   & 16.6  & 22.8  & 28.0  \\
    \hlineB{3}
    \multicolumn{9}{c}{The configuration of 1x Baseline can be found in the TFA official repo.}
    \end{tabular}
    }\label{tab:lvis-full}
\end{table}

\section{Discussion}

\subsection{Decoupling localization from classification.}
Consistent with the observation in~\cite{papyan2020prevalence}, by enhancing inter-class separation and intra-class compactness, the detection scores are improved. However, the features for localization should still be class-independent (\eg, bus and elephant has similar shape). 
From the implementation details, a projector is set where its input \& output are used for localization \& classification separately.
Then, sharing the features for localization and classification will lead to slight performance drop (\ie, AP$_{50}$ 74.0, nAP$_{50}$ 55.6).
As such, it is important to decouple the features for localization and classification and employing a simple linear projector has been shown to be userful.

\subsection{Design of Background class}

An object detector should reject the background and not recognize it as any foreground object. As such, a background class $\negcls{}$ is set as a placeholder and is trained to have high similarity with background proposals. Different from foreground objects, as background proposals can be diverse, we considered different strategies in designing the background class center. 

We first choose to separate the design of $W_b \cup W_n$ and $\mathbf{w}_{\negcls{}}$, \ie, deriving fixed offline weights for $\basecls{} \cup \novelcls{}$ only but learn the weight $\mathbf{w}_{\negcls{}}$. Then, we follow the open-set strategy~\cite{zhou2021learning} to set multiple background centers $W_-=\{\mathbf{w}_{\negcls{}}^{(i)}\}_{i=1}^{N_-}$ where $N_-$ is the number of background centers where the maximum logit, \ie, $\text{max}_{1\leq i \leq N_-}(\mathbf{x}^T \mathbf{w}_{\negcls{}}^{(i)})$, is used in classification. As compared in Table~\ref{tab:background}, having more learnable class centers can introduce trivial performance improvement but will drop clearly when $N_-$ is too large. However, when we directly set the classifier for the all classes ,\ie, $W_n \cup W_b \cup W_-$ as ETF , the performance drops when $N_- > 1$.

In practice, we observe all learnable negative weights $W_-$ are trained to separate from the $W_b \cup W_n$ where the weights in $W_-$ are still close to each other such that the diversity of background features are preserved indirectly. Instead, having all negative weights maximally separated from each other assume background features is very diverse and make the model hard to learn. As such, we choose to set $N_-=1$ and adjust margins through self-distillation to maintain the diversity properly.

\begin{table}[]
    \centering
    \renewcommand\thetable{M3}
    \caption{Ablation study of Background Design.}
    {
        \begin{tabular}{ccc|cccc}
        \hlineB{3}
            Idx & $N_{-}$ & Fixed & AP$_{50}$   & bAP$_{50}$  & nAP$_{50}$ \\ \hline
            1 & 1 & \checkmark{}  & 74.9 & 81.0  & 56.4  \\
            2 & 5 & \checkmark{}  & 73.5 & 81.3  & 50.2  \\
            3 & 1 &  & 74.9 & 81.0  & 56.4  \\
            4 & 5 &  & 74.9 & 80.9  & 56.7  \\
            5 & 10&  & 74.9 & 80.9  & 57.0  \\
            6 & 20&  & 74.6 & 81.1  & 55.2  \\
        \hlineB{3}
        \end{tabular}
    }\label{tab:background}
\end{table}

\subsection{More Visualization}

\begin{figure*}[!ht]
    \centering
    \includegraphics[width=0.8\textwidth]{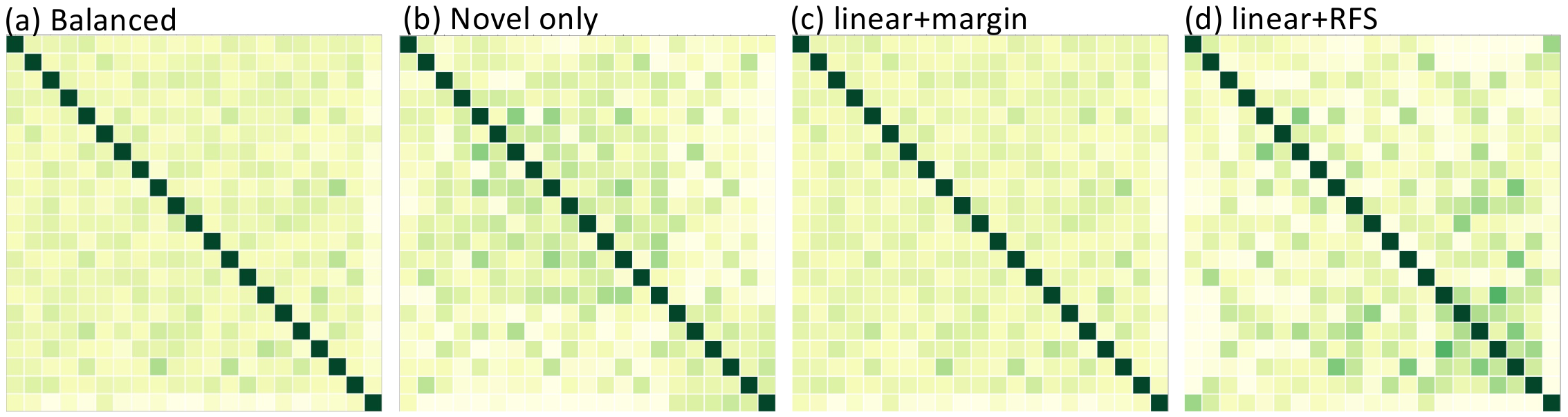}
    \renewcommand\thefigure{M1}
    \caption{Visualization classifier centers.}
    \label{fig:supp-fig}
\end{figure*}

As shown in Fig.~\ref{fig:supp-fig}, we visualize the classifier centers by their pair-wise cosine similarity when they are learned from scratch. Fig.~\ref{fig:supp-fig}(a) is the same as the Fig. 3(b) in the main paper but the background class center is also included (the rightmost and the bottom one). We can then see that when we have both base and novel annotation in the train set, the class centers can be trained to distance from all of the background classes. However, when we only use novel classes during the adaptation stage (Fig.~\ref{fig:supp-fig}(b)), the negative class center can be close to the novel class centers. Meanwhile, when we use the full set for training from scratch, we can see that the applying either RFS or adding margins can help with separating the novel class centers from the background class centers, while adding margins is more important.

The foreground class names (sorted by decreasing order) are person, chair, car, bottle, dog, potted plant, cat, boat, sheep, aeroplane, bicycle, tv monitor, horse, dining table, train, motorbike, cow, bus, bird, sofa.

\newpage

\section{Comparison with related work}

In this section, we provide in-detail comparison with a few representation works to highlight our motivation and contribution. All of the approached listed below have been briefly mentioned in the sections of Related Work and Experiment.

\paragraph{CME}~\cite{li2021beyond} similarly employs a margin equilibrium strategy in the few-shot finetuning. The determination of the margin value is based on the degree of feature disturbance which is measured by the scale of gradient among augmented samples.
Meanwhile, CME is motivated by the trade off between margins of base classes and the variance of novel classes. 

However, we have used the geometric property of Simplex ETF to maximally separate the feature clusters for all classes. In this way, we decouple the learning for inter-class separation and intra-class compactness and only tighten the feature cluster to the corresponding class centers to reach a balanced distribution. As such, we can learn discriminative features for all of the classes even on an extremely imbalanced dataset.

In addition, CME is still trained on the balanced dataset $\basesubset{} \cup \novelset{}$ and the so-called margin equilibrium is realized when the model is trained on a balanced set. Thus, CME may still forget the base knowledge. 
Instead, our margins are for all classes based on the prior of instance distribution and our approach is orthogonal to CME. Furthermore, the margin estimation strategy in CME can be used as an alternative of our self-distillation in margin adjustment.

\paragraph{Negative margin on few-shot classification} is studied in \cite{liu2020negative} and reveals the trade-off of classification accuracy between base recognition and novel  recognition. Namely, for a feature extractor pre-trained on \textit{base} classes, if the model achieves better test accuracy on the base classification, the adaptation accuracy towards classification accuracy is then minimized. As such, a comprehensive study is provided in \cite{liu2020negative}.

In contrast, we focus on few-shot object detection and aims to improve the few-shot adaptation efficiency without scarifying the performance of base detection. We always add positive class-specific margins to all classes where the margin values are adaptively learned during network training.

\paragraph{LOCE~\cite{feng2021exploring}} is applied on long-tail object detection, which is a more general case of generalized few-shot object detection (\ie, in GFSOD, the imbalance between base set and novel set is more significant and thus more challenging). A common problems discussed in LOCE and our paper is that the instance distribution of classes cannot be directly used to estimate the margins. 

As such, LOCE discards the prior and introduce the Equilibrium loss to use the mean classification score to determine the margin. In addition, they proposed a complex memory-augmented feature sampling to facilitate the network training.
In contrast, we clearly discuss and decouple the training objective for inter-class separation and intra-class compactness. 

We consider the distribution of classifier weights in conventional training and use ETF as a fixed classifier. In this case, we used the assigned weights to guide the separation of feature clusters between different classes, and then apply different margins to push the features to the assigned centers. As we apply margins to facilitate the balanced distribution, we can use the instance distribution as prior and use a simple knowledge distillation to adjust the margins and facilitate training.

\paragraph{Margin modification techniques} such as BALMS~\cite{ren2020balanced} and Seesaw loss\cite{wang2021seesaw} has been proposed. Specifically, BALMS considers the boundary shifting problem in long-tailed classification/segmentation and present a meta-sampling strategy to re-estiamte the boundary indicated in the Softmax function. Seasaw loss defines a compensation factor in vanilla cross entropy loss to balance the error for different classes. In both case, they in effect count on the real-time (online) distribution of selected samples during the training and then adjust the loss. Instead, we focus on the inter-class separation and intra-class compactness to guide the training of features, \ie, re-arranging the feature distribution from the perspective of feature geometry. In addition, the margin modification techniques can be used as an alternative of our margin adjustment strategy for the intra-class compactness only.


\paragraph{Connection with FSCE} In FSCE~\cite{sun2021fsce}, the authors has provided a strong baseline by adjusting the hyper-parameters in RPN and proposal selection. We have tried to apply it in our framework but the performance drops. As such, we still follow the hyper-paramter setting in TFA. Meanwhile, it also demonstrates that the observation in FSCE is only available in the two-step based training strategy such as TFA, and cannot be generalized to a universal case.

Furthermore, FSCE proposed a contrastive encoding approach and treats the proposals as augmentation of the same instance. However, we have also add the contrastive loss in our approach and observed that it may help improve the nove detection slightly but hurt the base detection significantly. We think the reason is that the data distribution is extremely imbalanced and and the contrastive loss cannot help.


{\small
\bibliographystyle{cvpr2023/ieee_fullname}
\bibliography{cvpr2023/egbib}
}

\end{document}